%% file: templateArxiv.tex
\documentclass{article}

\usepackage{PRIMEarxiv}

\usepackage[utf8]{inputenc} 
\usepackage[T1]{fontenc}    
\usepackage{hyperref}       
\usepackage{url}            
\usepackage{booktabs}       
\usepackage{amsfonts}       
\usepackage{nicefrac}       
\usepackage{microtype}      
\usepackage{lipsum}
\usepackage{multirow}
\usepackage{fancyhdr}       
\usepackage{graphicx}       
\graphicspath{{media/}}     
\AtBeginDocument{%
  }
\usepackage[most]{tcolorbox}   
\newtcblisting[auto counter]{promptbox}[2][]{%
float,                                   
  floatplacement=ht,
    toptitle=0pt,      
  bottomtitle=0pt,   
  colback=gray!5,
  colframe=gray!40,
  fonttitle=\bfseries,
  title={Prompt~\thetcbcounter: #2},
  listing only,                            
  listing options={
    basicstyle=\ttfamily\tiny,
    breaklines=true,                       
    columns=fullflexible,
    keepspaces=true,                       
    escapeinside={(*@}{@*)},
    upquote=true                           
  },
  top=0pt,bottom=0pt,left=0pt,right=0pt,
    before upper=\vspace{0pt},    
  after upper=\vspace{0pt},     
  label={#1},
}
\title{Are Large Language Models Dynamic Treatment Planners? An In Silico Study from a Prior Knowledge Injection Angle
}
\author{
  Zhiyao Luo\\
  Department of Engineering Science \\
  University of Oxford \\
  \texttt{zhiyao.luo@eng.ox.ac.uk}
  \And
  Tingting Zhu \\
  Department of Engineering Science \\
  University of Oxford \\
  \texttt{tingting.zhu@eng.ox.ac.uk}
}

\begin{document}
\maketitle
\begingroup
\renewcommand\thefootnote{}\footnotetext{Preprint.}\addtocounter{footnote}{-1}
\endgroup

\begin{abstract}
Reinforcement learning (RL)-based dynamic treatment regimes (DTRs) hold promise for automating complex clinical decision-making, yet their practical deployment remains hindered by the intensive engineering required to inject clinical knowledge and ensure patient safety. Recent advancements in large language models (LLMs) suggest a complementary approach, where implicit prior knowledge and clinical heuristics are naturally embedded through linguistic prompts without requiring environment-specific training. In this study, we rigorously evaluate open-source LLMs as dynamic insulin dosing agents in an in silico Type 1 diabetes simulator, comparing their zero-shot inference performance against small neural network-based RL agents (SRAs) explicitly trained for the task. Our results indicate that carefully designed zero-shot prompts enable smaller LLMs (e.g., Qwen2.5-7B) to achieve comparable or superior clinical performance relative to extensively trained SRAs, particularly in stable patient cohorts. However, LLMs exhibit notable limitations, such as overly aggressive insulin dosing when prompted with chain-of-thought (CoT) reasoning, highlighting critical failure modes including arithmetic hallucination, temporal misinterpretation, and inconsistent clinical logic. Incorporating explicit reasoning about latent clinical states (e.g., meals) yielded minimal performance gains, underscoring the current model's limitations in capturing complex, hidden physiological dynamics solely through textual inference. Our findings advocate for cautious yet optimistic integration of LLMs into clinical workflows, emphasising the necessity of targeted prompt engineering, careful validation, and potentially hybrid approaches that combine linguistic reasoning with structured physiological modelling to achieve safe, robust, and clinically effective decision-support systems.
\end{abstract}


\maketitle

\section{Introduction}
The integration of reinforcement learning (RL) into dynamic treatment regimes (DTRs) has showcased considerable promise for automating complex clinical decisions, particularly glycaemic control, in carefully controlled settings~\cite{zhu2020insulin}. However, the practical implementation of RL approaches faces significant challenges, including the need for carefully engineered reward signals, observation space, and action space. To achieve robustness against pharmacokinetic/pharmacodynamic variability, developers must inject sufficient prior knowledge during training, such as policy safety constraints\cite {cutler2015efficient}, world modelling~\cite{tirumala2022behavior} or direct human advice~\cite{judah2010reinforcement}. However, existing methods are mostly from robotics areas, where adding human demonstrations or applying constraints is at a lower cost. In the field of treatment regimes, there is no silver bullet to inject prior knowledge before or during training.

Recent advances in large language models (LLMs) present an alternative and complementary approach that could mitigate these limitations~\cite{bedi2024systematic, wang2025survey}. LLMs possess distinctive capabilities that make them promising candidates for clinical decision support. First, their proficiency in in-context learning and instruction-following facilitates embedding straightforward prior knowledge, which is typically challenging to encode explicitly in smaller neural network models~\cite{an2024make}. Second, their pre-trained medical knowledge may enhance the quality and safety of clinical recommendations. LLMs also exhibit sophisticated reasoning capabilities, especially when analysing complex historical EHR data~\cite{shi2024ehragent}. The potential of LLMs for generating treatment recommendations remains largely unexplored, although previous research has extensively explored the reasoning abilities of LLMs across domains such as mathematics, coding, robotics, and medical diagnosis~\cite{plaat2024reasoning}. LLMs can potentially capture the rich temporal dependencies inherent in patient trajectories, provide human-readable rationales, and offer safer, more interpretable treatment policies without explicit reward optimisation.

This work investigates whether open-sourced state-of-the-art LLMs can generate clinically meaningful insulin dosing recommendations purely through inference. We directly compare their performance with small neural-network-based RL agents within the same experimental framework. Specifically, our investigation aims at exploring four key research questions (RQs). \textbf{(RQ1)} How can expert clinical knowledge be effectively injected into RL, and can LLMs simplify this process? \textbf{(RQ2)} Can LLMs outperform small neural-network-based RL policies in zero-shot inference settings? \textbf{(RQ3)} Do the parameter scaling law and sampling temperature findings observed in other LLM tasks extend to dynamic treatment regime scenarios? \textbf{(RQ4)} Does incorporating chain-of-thought (CoT) prompting improve reasoning quality and clinical performance in dynamic treatment scenarios? To analyse these questions, we use the SimGlucose type-1 diabetes simulator~\cite{man2014uva}, a validated in silico environment that rigorously evaluates insulin dosing policies.

\section{Problem Formulation and Environment Setting}
\label{sec:llm-problem formulation}
We use the SimGlucose environment~\cite{luo2024dtr} to validate the capabilities of LLMs for two critical considerations. First, despite having a seemingly simple observational structure, consisting primarily of blood glucose measurements, the environment exhibits highly complex underlying dynamics. As demonstrated in previous work~\cite{luo2024dtr}, maintaining glycaemic control requires precise dosage adjustments, where even advanced off-policy RL algorithms struggle under realistic noise and inter-patient pharmacokinetic/pharmacodynamic variability. Second, the simulator presents straightforward rules, such as "insulin reduces glucose", which are challenging to embed explicitly into traditional RL policies. In contrast, LLMs can directly incorporate prior knowledge through natural language instructions, making the SimGlucose environment an ideal testbed to assess whether LLMs excel as clinicians who follow instructions. 

The patient's state is observed in a discrete timestep of 15 minutes and includes the preceding four-hour glucose trajectory and the insulin infusion history. The agent selects a continuous (for PPO~\cite{schulman2017proximal}) or 11-bin discrete (for DQN~\cite{mnih2015human}) insulin infusion rate $a_t \in [0,9]$ U/h, administered as a basal rate over the subsequent 15-minute control interval. The clinical goal is to maintain blood glucose within the target range $[70,140]$ mg/dL, avoiding severe excursions below 40 mg/dL or above 500 mg/dL. Episodes end after 16 hours (64 time steps) or immediately after glucose excursions above $(40,500)$ mg / dL.

\section{Reward and Evaluation Protocol}
\label{sec:llm-Reward and Evaluation Protocol}
To rigorously evaluate insulin dosing strategies generated by small reinforced models and LLMs, we adopt three key evaluation metrics: 

\begin{itemize}
    \item \textbf{Survival Rate:} Defined as the proportion of episodes that complete the full 16-hour simulation (64 timesteps) without termination due to extreme blood glucose (BG) excursions. Termination occurs if BG drops below 40 mg/dL or exceeds 500 mg/dL.

    \item \textbf{Time-In-Range (TIR):} A standard metric that quantifies the fraction of time steps in which BG lies within the recommended range of 70–180 mg/dL. This assesses the 'control quality' and is computed as the average hit rate across the episode.
 
    \item \textbf{Normalised Return:} Derived from the total episodic return specified in the reward design (See Section~\ref{app:sec-state-action-reward} in Appendix, this metric captures the cumulative efficacy of treatment using the glucose risk index. We min-max scale the return to the $[0, 100]\%$ range using:$\frac{\text{Return}- (- 99.7)}{64 -(- 99.7)} \times 100$, where -99.7 is the minimum possible return and 64 is the maximum (survival with optimal glycaemic control).
\end{itemize}
All metrics are computed over three groups of four patients: adults, adolescents, and children. In SimGlucoseEnv, adult physiology is modelled as the least volatile, and paediatric physiology as the most volatile. Consequently, cohorts can be interpreted as representing increasing levels of difficulty: easy (adult), medium (adolescent), and hard (child). Each patient undergoes 20 test episodes (4 seeds × 5 repeats), delivering sufficient statistical power. We report metrics using bootstrapped confidence intervals. The error bars represent the 95\% confidence interval derived by bootstrapping each metric 1,000 times with replacement and selecting the 2.5\% and 97.5\% percentiles, respectively. The primary reported 'mean' metric is the mean of the bootstrapped result. The notation is used for all the following results, unless otherwise specified.

For small reinforced agents (SRAs), DQN and PPO are trained in adult, adolescent, child and mixed environments, respectively, and the best model is selected by achieving the best overall evaluation return. The best selected model is then evaluated on each stratum in the evaluation environments for generalisation assessment.

For LLMs, the pretrained language models are evaluated directly in the evaluation environments with different prompt methods, which will be introduced in Section~\ref{sec:llm-LLMs in Insulin Dosing}.

\section{Experimental Setting}
\subsection{Small Reinforced Agents in Insulin Dosing}
\subsubsection{Training Small Neural Networks with Reinforcement Learning}
We selected Deep Q-Network (DQN) and Proximal Policy Optimisation (PPO) as representative algorithms for off-policy and on-policy reinforcement learning, respectively. Both algorithms were trained separately under four distinct scenarios to account for varying levels of patient variability: adult only (easy), adolescent only (medium), child only (hard) and a mixed group from all three categories of patients. Each training epoch consisted of 480 simulation steps, equivalent to 10 full episodes without early termination. Training was performed over 20 epochs, with model checkpoints saved and evaluated at the end of each epoch in their respective training environments. The best-performing checkpoints were selected based on the highest normalised episodic training return, and were subsequently evaluated on the complete evaluation environment set described in Section~\ref{sec:llm-problem formulation}. Detailed hyperparameter configurations and training specifics are provided in Appendix~\ref{app:llm-sra-hyperparameters}.

\subsubsection{Injecting Prior Knowledge in Small Reinforced Agents via Exploration}
Integrating clinical prior knowledge directly into RL poses challenges due to the difficulty neural networks have in interpreting natural language or expert intuition directly. Due to the sparse dosage nature of DTR, we choose to incorporate clinical insights into the exploration strategy, thus guiding clinically plausible actions from the early training stages. 

For DQN, exploration typically employs an $\varepsilon$-greedy strategy, where actions are selected uniformly throughout the action space. In insulin dosing scenarios, this approach often leads to unsafe, clinically implausible actions (e.g., overdosing), as the optimal insulin dosage is frequently zero. To mitigate this issue, we introduce a domain-informed modification that assigns the zero-dose action a significantly higher exploration probability. Denoting the probability of selecting action 0 (i.e., zero dose) as $p_0$ and selecting any other action as $p_j, j\in[1, 2, ...,|A| -1]$, we design a heuristic
$$
\frac{p_0}{p_j} \;=\; \frac{|A|(|A|+1)}{|A|-1}
$$
where $|A|$ is the number of discrete dose levels. Since we use an 11-bin action space for DQN, the probability of selecting zero dosage during exploration is around 0.569. Further justifications of the heuristic design can be found in \ref{app:llm-heuristic}

For PPO, the integration of prior knowledge differs due to continuous action spaces. We explored two primary action transformations: a smooth hyperbolic tangent (Tanh) function and a hard-clipping function. The clip transformation is commonly used due to simplicity and stability, and is considered the `without prior knowledge' choice. In contrast, the Tanh transformation (considered the `with prior knowledge' choice) provides smooth action saturation, implicitly embedding prior knowledge that favours conservative dosing decisions unless strongly justified by model predictions. Detailed theoretical analyses and justifications comparing these transformations are provided in Appendix~\ref{app:llm-tanh}. 

In addition, we initialise the PPO actor network with near-zero weights and zero biases to ensure that, at the beginning of training, the policy outputs actions close to zero. This encourages safe, low-dose decisions during the early learning phase. To further align initial exploration with prior knowledge, we warm-start the training buffer using a constrained random policy biased toward small doses. For PPO, actions are uniformly sampled from the continuous interval [0,2] U/h. For DQN, actions are sampled uniformly from the discrete set \{0, 1, 2\}, where actions 0,1,2 correspond to dose 0, 0.9 and 1.8 units/hour. This strategy ensures that early-stage trajectories observed by the agent predominantly reflect conservative dosing behaviours, thereby promoting safer exploration from the outset.

\subsection{ LLMs in Insulin Dosing}
\label{sec:llm-LLMs in Insulin Dosing}
\subsubsection{Basic LLM Policy Construction}
To facilitate LLM inference on structured medical time series, we first transformed numerical patient-state representations into descriptive text. Specifically, each observation was formatted to include the patient's glucose measurements, insulin infusion rates, and corresponding timestamps at 15-minute intervals. This structured format mirrors real-world EHRs and leverages previous research demonstrating improved LLM reasoning when inputs are presented as formatted text~\cite{kojima2022large, wei2022emergent}. An illustrative example of a four-hour glucose trajectory (covering 16 discrete timesteps) is provided in Prompt~\ref{prompt:observation-example}. 

\begin{promptbox}[prompt:observation-example]{Observation Example}
Day 1, Time: 05:00:00 (initial measurement), glucose: 159.27 mg\/dL, insulin rate: 0.0000 unit\/hour, insulin dose: 0.00 unit.

Day 1, Time: 05:15:00, glucose: 148.96 mg\/dL, insulin rate: 1.0000 unit\/hour, insulin dose: 0.25 unit.

Day 1, Time: 05:30:00, glucose: 149.71 mg\/dL, insulin rate: 0.0000 unit\/hour, insulin dose: 0.00 unit.

...
\end{promptbox}
\begin{promptbox}[prompt:base-sys]{Basic Zero-shot Prompt Template}
You are a clinical specialist responsible for managing patients with Type-1 Diabetes. Your primary objective is to maintain the patient's blood glucose levels within the safe range of **70-140 mg/dL** by administering appropriate insulin doses.

### Monitoring and Decision Frequency
Blood glucose levels are observed every **15 minutes**. Your task is to determine the insulin dose rate in units/hour every **15 minutes** based on the latest glucose readings and trends.

### Insulin Administration
    - Insulin lowers blood glucose levels with a delayed effect. Your decisions will define the insulin dose rate to be administered over the next **15 minutes** to maintain a normal glucose level in a long run.
    
    - We do not distinguish between basal and bolus insulin. The dosing decision is based on the latest glucose readings and trends.
    
    - **Rate Range**: You MUST provide a dose in the range of [0 to 9] units/hour, inclusive. 
    
    - **Administration Interval**: The specified dose is distributed evenly over the 15-minute period, i.e., the total dosage is your_action/60*15. You only need to provide the dose rate in units/hour. Do NOT specify the total dose.
    
###Observations
(*@\textcolor{blue}{<Observation>}@*)

###Request
Determine the optimal insulin rate for the current 15-minute interval to maintain a patient's blood glucose levels within the safe range of 70-140 mg/dL. Choose a dosage value. For example, if you choose 0 units, enter 0. DO NOT say anything else. 

###Answer
\end{promptbox}

The baseline LLM policy was constructed using a straightforward system prompt to maintain blood glucose within the safe range of 70–140 mg/dL by selecting an appropriate insulin infusion rate every 15 minutes. This prompt explicitly specifies both the clinical goals and the action constraints, including limiting insulin dosage rates to the clinically acceptable range of 0-9 units/hour. A minimal instruction was added to request a numeric response, forming the zero-shot inference policy prompt (see Prompt~\ref{prompt:base-sys}). This minimalist, zero-shot prompting strategy served as the foundational baseline for all subsequent LLM policy evaluations, based on prior evidence that even minimal natural language supervision can effectively guide LLMs in decision-making scenarios.

\subsubsection{Injecting Prior Knowledge in LLMs}

\begin{promptbox}[prompt:full-sys]{Expert Knowledge System Prompt}
(*@\textcolor{blue}{<Same base system prompt as Prompt~\ref{prompt:base-sys}>}@*)

### Hidden Variables
- **Food Intake**: Food consumption increases blood glucose levels.
- **Exercise**: Exercise reduces blood glucose levels.
- **Estimation**: Since food intake and exercise are not directly observable, estimate based on time of day and observed glucose trends using clinical judgment and common sense.

### Penalties and Risks
- **Blood Glucose Outside Safe Range (70-140 mg/dL)**:
  - **Above 140 mg/dL**: Hyperglycaemia penalties.
  - **Below 70 mg/dL**: Hypoglycaemia penalties, with increased severity.
  - **Above 500 or below 40 mg/dL**: EXTREMELY DANGEROUS, your treatment will be considered a failure and the patient will die!
- **Insulin Dose Considerations**:
  - **High Doses**: Use cautiously to avoid rapid and excessive lowering of glucose levels.
  - **Low Glucose Levels (<70 mg/dL)**:
    - **Action**: Immediately cease insulin administration until glucose levels rise above 70 mg/dL.
    - **Priority**: Prevent hypoglycaemia due to its acute dangers.

### Safety Precautions
- **Avoid Overdosing Insulin**: Prevent hypoglycaemia by carefully balancing insulin doses.
- **Insulin Stacking Awareness**: You should consider the accumulated dosage and the delayed effect of insulin on glucose levels carefully.
- **Prioritize Patient Safety**: Always aim to keep glucose levels within the target range. If uncertainty exists, opt for a lower or zero insulin dose to ensure safety.

###Observations
(*@\textcolor{blue}{<Observation>}@*)

###Request
Determine the optimal insulin rate for the current 15-minute interval to maintain a patient's blood glucose levels within the safe range of 70-140 mg/dL. Choose a dosage value. For example, if you choose 0 units, enter 0. DO NOT say anything else. 

###Answer
\end{promptbox}

\begin{promptbox}[prompt:cot-instruction]{Zero-shot CoT Instruction}
Determine the optimal insulin rate for the current 15-minute interval to maintain a patient's blood glucose levels within the safe range of 70-140 mg/dL. First, analyse the current state step-by-step. Finally, you must choose a dosage value enclosed in answer tags (i.e., <ans> and </ans>]), for example, <ans>0</ans>, without any non-numerical word. Let's think step by step.   
\end{promptbox}

\begin{promptbox}[prompt:meal-cot-instruction]{Zero-shot CoT With Meal Information Instruction}
Determine the optimal insulin rate for the current 15-minute interval to maintain a patient's blood glucose levels within the safe range of 70-140 mg/dL. First, analyse the current state step-by-step by estimating the Correction Factor(CF) and Total Daily Insulin(TDI). If the patient has taken any meal, you must consider the meal effect. Finally, you must choose a dosage value enclosed in answer tags (i.e., <ans> and </ans>]), for example, <ans>0</ans>, without any non-numerical word. Let's think step by step.
\end{promptbox}

To add priors to LLMs, we extend the baseline prompt by embedding prior knowledge directly into the system instruction (Prompt~\ref{prompt:full-sys}). This expert-augmented prompt includes additional guidance, such as the delayed effect of insulin, safety thresholds for hypoglycaemia, and cautionary principles around high-dose administration. It also introduces reasoning about hidden variables, such as estimating food intake or exercise effects from indirect indicators like time of day or rising glucose trends. By explicitly encoding these safety rules and latent clinical patterns, our objective is to improve the plausibility and robustness of the dosing decisions of the model. In recent work, similar approaches have been explored that aim to ground LLM reasoning in clinical heuristics~\cite{hager2024evaluation}. We then evaluate whether language models can engage in explicit clinical reasoning using the zero-shot CoT prompting strategy (Using the instruction in Prompt~\ref{prompt:cot-instruction} and the same system prompt as Prompt~\ref{prompt:full-sys}). Here, the instruction requires the model to analyse the current patient's condition step by step before selecting a dose. The final answer is wrapped in delimiters (<ans>answer</ans>) to isolate the scalar output from the reasoning trace. Lastly, we explore a variant of chain-of-thought prompting that attempts to uncover latent clinical variables, particularly the presence of meals and the patient’s insulin sensitivity (Using instruction Prompt~\ref{prompt:meal-cot-instruction} and using the same system prompt as Prompt~\ref{prompt:full-sys}). The prompt encourages the model to estimate factors such as total daily insulin (TDI) or correction factor (CF), and to infer hidden states from glucose dynamics when explicit indicators are missing. This is particularly relevant in partially observed settings like SimGlucose, where meal events are recorded as binary tokens, but not described explicitly in natural language. By embedding instructions to reason over such hidden variables, we assess whether LLMs can estimate hidden variables and the environmental PK/PD dynamics. Across all types of prompts, LLMs act solely via inference without training.

\section{Results}
We present an extensive evaluation comparing LLMs with SRAs. Specifically, we examine two representative open-source LLM families, Qwen2.5~\cite{team2024qwen2} and LLaMA3~\cite{grattafiori2024llama}, selected due to their robust general-purpose reasoning capabilities and widespread adoption in the community.

\subsection{The effectiveness of Prior Knowledge Injection on SRAs}

\begin{figure}[ht]
    \centering
    \includegraphics[width=\linewidth]{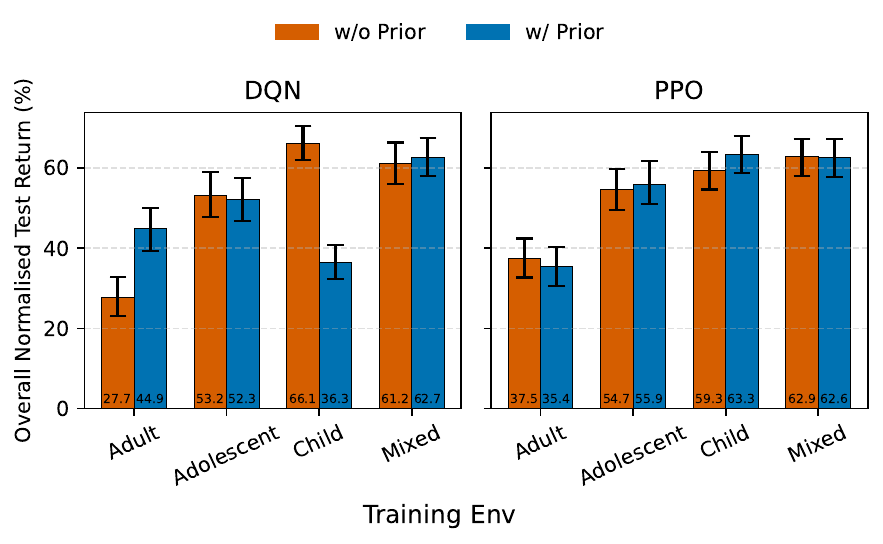}
    \caption[Overall normalised return of SRA on the evaluation set for DQN and PPO.]{Overall normalised return of SRA on the evaluation set for DQN and PPO. Each algorithm is trained on one of four environment types (adult, adolescent, child, or mixed cohort) and evaluated across all environment types. The reported results represent the overall average performance across adult, adolescent and child environments. Evaluation result for each strata is not shown here for clarity, and will be discussed later in Figure \ref{fig:LLMs-SRAs}. `Prior' denotes prior knowledge, `w/' means 'with' and `w/o' means `without'.}
    \label{fig:llm-small-model-summary}
\end{figure}
    
\begin{table}[ht]
\small
    \centering
        \begin{tabular}{@{}lcccc@{}}
            \toprule
            \multirow{2}{*}{Train Env.} & \multicolumn{2}{c}{\textbf{DQN}} & \multicolumn{2}{c}{\textbf{PPO}} \\ \cmidrule(lr){2-3} \cmidrule(l){4-5}
             & w/o  $\rightarrow$ w/ Prior & $\Delta\,(\%)$ & w/o  $\rightarrow$ w/ Prior & $\Delta\,(\%)$ \\ \midrule
            Adult          & 27.7 $\rightarrow$ 44.9 &  $\phantom{-}$17.2 & 37.5 $\rightarrow$ 35.4 & $-2.1$ \\
            Adolescent     & 53.2 $\rightarrow$ 52.3 & $-0.9$           & 54.7 $\rightarrow$ 55.9 &  $\phantom{-}$1.2 \\
            Child          & \textbf{66.1} $\rightarrow$ 36.3 & $-29.8$          & 59.3 $\rightarrow$ 63.3 &  $\phantom{-}$4.0 \\
            Mixed          & 61.2 $\rightarrow$ 62.7 &  $\phantom{-}$1.5 & 62.9 $\rightarrow$ 62.6 & $-0.3$ \\ \bottomrule
        \end{tabular}%
    \caption[Impact of prior knowledge injection on overall normalised episodic return on the evaluation environments.]{Impact of prior knowledge injection on overall normalised episodic return on the evaluation environments. For clarity, this table only shows the mean value captured from Figure~\ref{fig:llm-small-model-summary}. $\Delta$ represents the performance gain or loss by percentage. `Prior' denotes prior knowledge, `w/' means 'with' and `w/o' means `without'.}
    \label{tab:llm-prior-impact}
\end{table}

Figure~\ref{fig:llm-small-model-summary} and Table~\ref{tab:llm-prior-impact} contrast the normalised episodic returns obtained by DQN and PPO trained with and without explicit prior knowledge across four different training environments. The effects of embedding prior knowledge into DQN and PPO appear contradictory. For DQN, the introduction of prior knowledge significantly increases performance in the easiest (adult) environment, improving from 27.7\% to 44.9\%. However, this benefit diminishes sharply with increasing difficulty, becoming negative in the hardest (child) environment, where performance declines markedly from 66.1\% to 36.3\%. In contrast, PPO demonstrates modest performance improvements upon the introduction of prior knowledge, particularly in harder environments; returns slightly increase from 54.7\% to 55.9\% in the medium environment and more notably from 59.3\% to 63.3\% in the child environment. In the mixed environment, neither algorithm shows a statistically significant difference, indicating that embedding explicit knowledge has a negligible marginal impact when training encompasses all levels of difficulty.

Interpreting the aforementioned results, the contradictory behaviour can be explained by the fundamental differences between the learning mechanisms of DQN and PPO. DQN, as an off-policy and value-based algorithm, initially benefits from conservative prior knowledge in easier (such as Adult) environments, since adults' glucose levels can be well maintained by lower insulin dosage given their stable glucose-insulin dynamics. However, in more challenging environments (such as Children), the same conservative strategy can induce severe hyperglycaemic episodes due to insufficient doses, resulting in substantial penalties. Repetition of these negative experiences rapidly deteriorates the landscape of learnt values, dramatically reducing performance. PPO, as an on-policy actor-critic method, continuously samples trajectories that are aligned with the current policy. Since the conservative prior knowledge injected into PPO is via action scaling, the prior does not impose strict constraints on policy, but gently biases exploration towards conservative dosing. Consequently, PPO avoids the severe penalties that undermine DQN in more volatile environments. In mixed training environments, prior knowledge injection shows less significant improvement because training already incorporates both slow and fast glucose dynamics, inherently encouraging context-sensitive policy adaptation. 

The above findings highlight the algorithm-specific and task-dependent nature of embedding prior knowledge. With extensive experiments on adding or removing prior knowledge, DQN w/o prior training on the child environments stands out as the best SRA due to its superior overall performance on the evaluation environments, outperforming its variants with prior knowledge and all PPO SRAs. Analysis shows that blindly applying prior knowledge to SRAs can lead to counterproductive results and pose challenges when generalised to a wider range of patient groups. These observations naturally lead us to ask: \textit{Can LLMs address these challenges with less manual effort}?

\subsection{Comparison between LLMs and SRAs}

\begin{figure}
    \centering
    \includegraphics[width=0.99\linewidth]{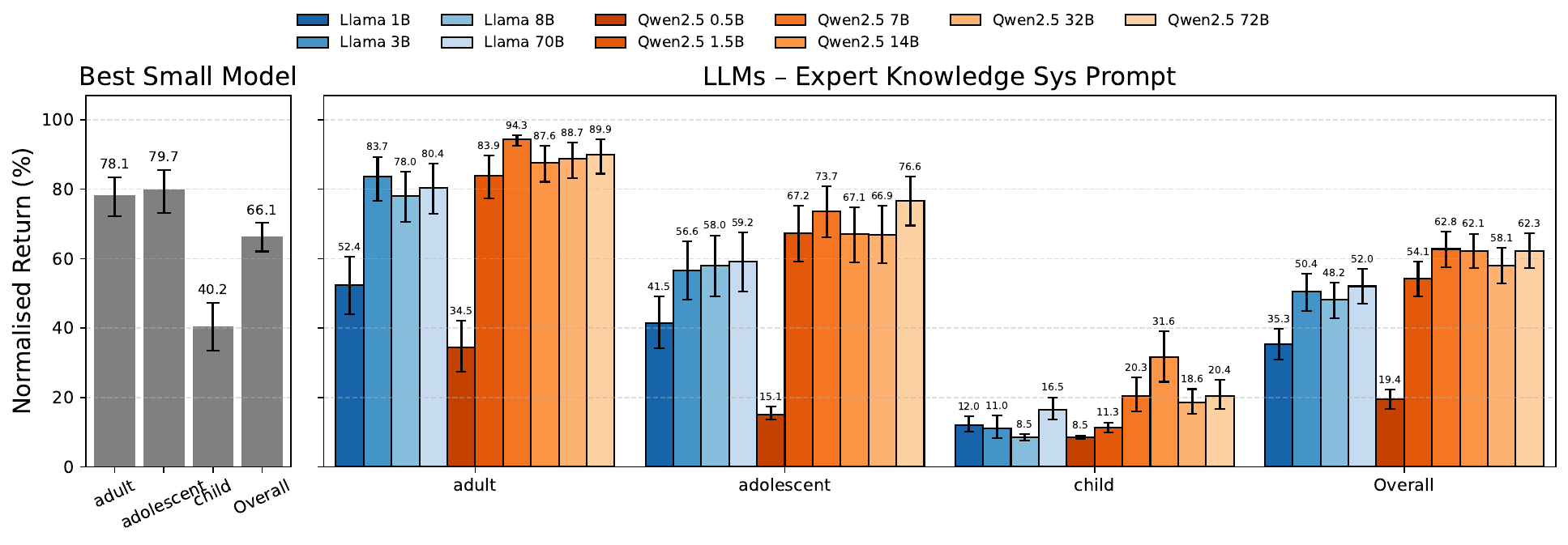}
    \caption[LLMs' normalised return performance compared with the best SRA model]{LLMs' normalised return performance compared with the best SRA model (i.e., DQN trained on child environments without prior knowledge) on each sub-evaluation task during testing. The blue colour spectrum represents the Llama3 series and the orange spectrum represents the Qwen2.5 series, respectively. A lighter colour denotes a larger parameter size (in billions).}
    \label{fig:LLMs-SRAs}
\end{figure}

The comparative evaluation between LLMs and the best SRA (i.e., DQN trained on `child' patient environments) reveals several nuanced insights, as shown in Figure \ref{fig:LLMs-SRAs}. Without any finetuning, certain LLMs can achieve performance on par with or even surpassing the best-trained SRAs. In terms of overall return, Qwen2.5 models with 7B, 14B, and 32B parameters achieved 62.8\%, 62.1\%, and 62.3\%, respectively, demonstrating the strong zero-shot capabilities of recent LLMs. The advantages of LLMs are primarily driven by superior performance in the adult cohort. In adult evaluation, both the Llama and Qwen2.5 models can overperform the best SRA (78.1\%) when the parameter size is greater than 1B. In contrast, across tasks for adolescents and children where physiological dynamics are more volatile, none of the LLM variants outperformed the best-performing SRA.

\subsection{Analysis of LLM Treatment Behaviour}
To better understand the mechanisms behind the clinical decision-making of LLMs, we examine how the model family, the parameter scale, the decoding temperature, and the prompt method influence the insulin dosing behaviour.

\paragraph{Impact of Model Family, Sampling Temperature and Parameter Scaling}
\begin{figure}[h]
    \centering
    \includegraphics[width=0.9\linewidth]{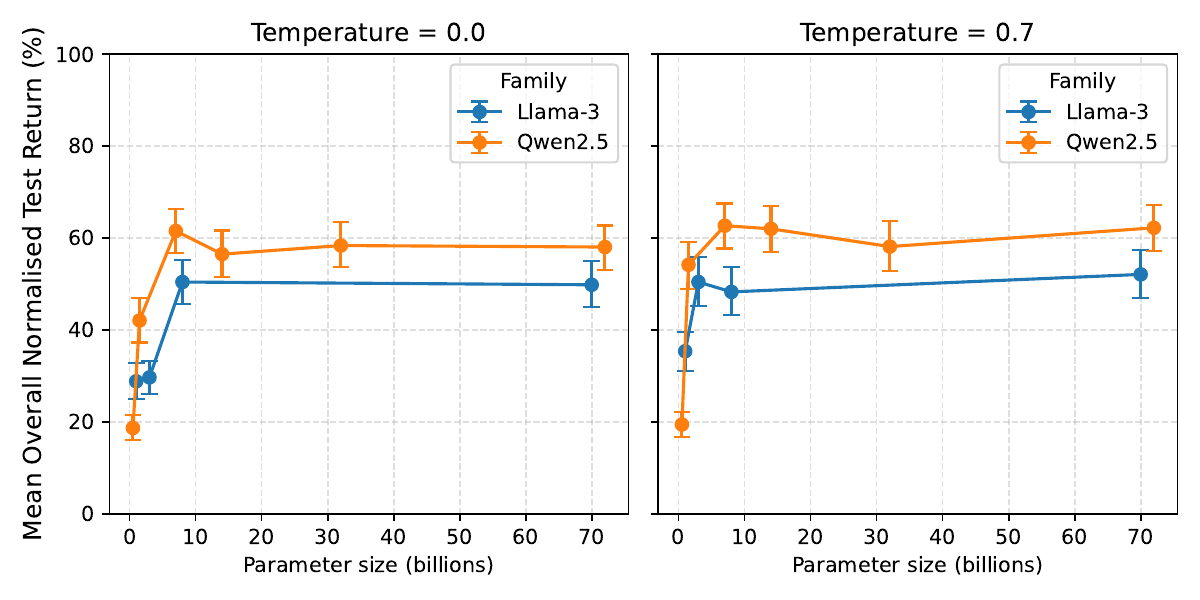}
    \caption{Comparison of insulin dosing performance across LLM families (Qwen2.5 and LLaMA3) for different temperature settings. }
    \label{fig:llm-family and temperature}
\end{figure}

We evaluated performance variations across different families and sizes of models, presented in Figure~\ref{fig:llm-family and temperature}. The Qwen2.5 model family consistently outperforms the LLaMA3 series across all parameter sizes evaluated, suggesting that intrinsic architectural differences or specific training methodologies significantly influence clinical decision-making capabilities. 

We compare deterministic decoding (temperature = 0) with stochastic decoding (temperature = 0.7), where a higher temperature introduces more randomness into the token sampling process. At temperature 0.7, the models consistently produce more diverse responses, yielding superior results compared to the more rigid and repetitive output generated at temperature 0.0. This aligns with observations from other LLM application domains~\cite{renze2024effect}. 

Clear parameter-scaling trends are observed on both Qwen2.5 and Llama from 1B to 7B, but no further improvement is observed from 7B to 70B. This finding is contradictory to common natural language tasks~\cite{isik2024scaling}, where the scaling law is typically a log-like curve.   

\paragraph{Influence of Prompting Strategies on Performance}
\begin{figure}[h]
    \centering
    \includegraphics[width=0.9\linewidth]{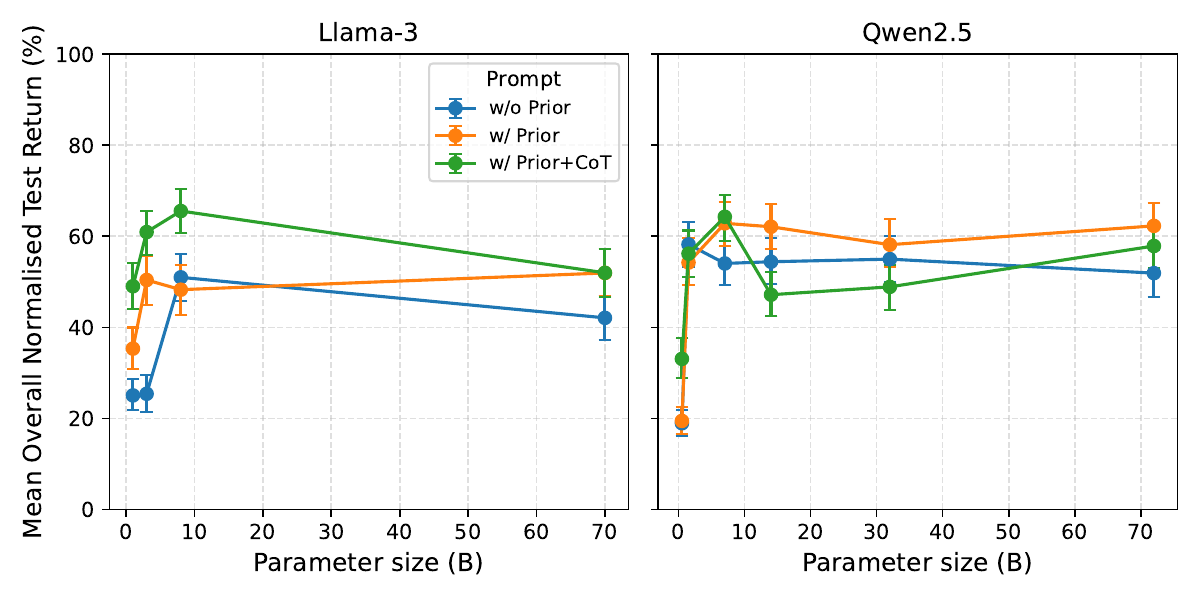}
    \caption{Impact of different prompting methods on insulin dosing for Qwen2.5 and LLaMA3 families for prompt settings: base zero-shot prompt (w/o prior), prior knowledge zero-shot prompt (w/ prior) and prior knowledge zero-shot-CoT prompt (w/ prior+CoT).}
    \label{fig:llm-prompt1}
\end{figure}

To derive a deeper observation of the behaviour of smaller LLMs versus their larger variants, we analyse the scaling trend under various prompt methods, as illustrated in Figure~\ref{fig:llm-prompt1}. Incorporating explicit prior knowledge (i.e., `w/ prior', in orange) into the prompt significantly improves the performance of the policy for both model families, with the improvement being particularly significant for Qwen2.5. Interestingly, chain-of-thought (CoT) with prior knowledge (i.e., 'w/ prior+CoT') prompting yields divergent results. For the LLaMA3 family, CoT with prior knowledge generally enhances treatment outcome, while for Qwen2.5, CoT with prior knowledge tends to degrade performance, especially in models exceeding 10B parameters. This unexpected finding challenges conventional wisdom drawn from general-purpose reasoning tasks, where CoT typically improves outcomes.

\paragraph{In-depth Analysis of Chain-of-Thought Prompting}
\begin{figure}
    \centering
    \includegraphics[width=0.9\linewidth]{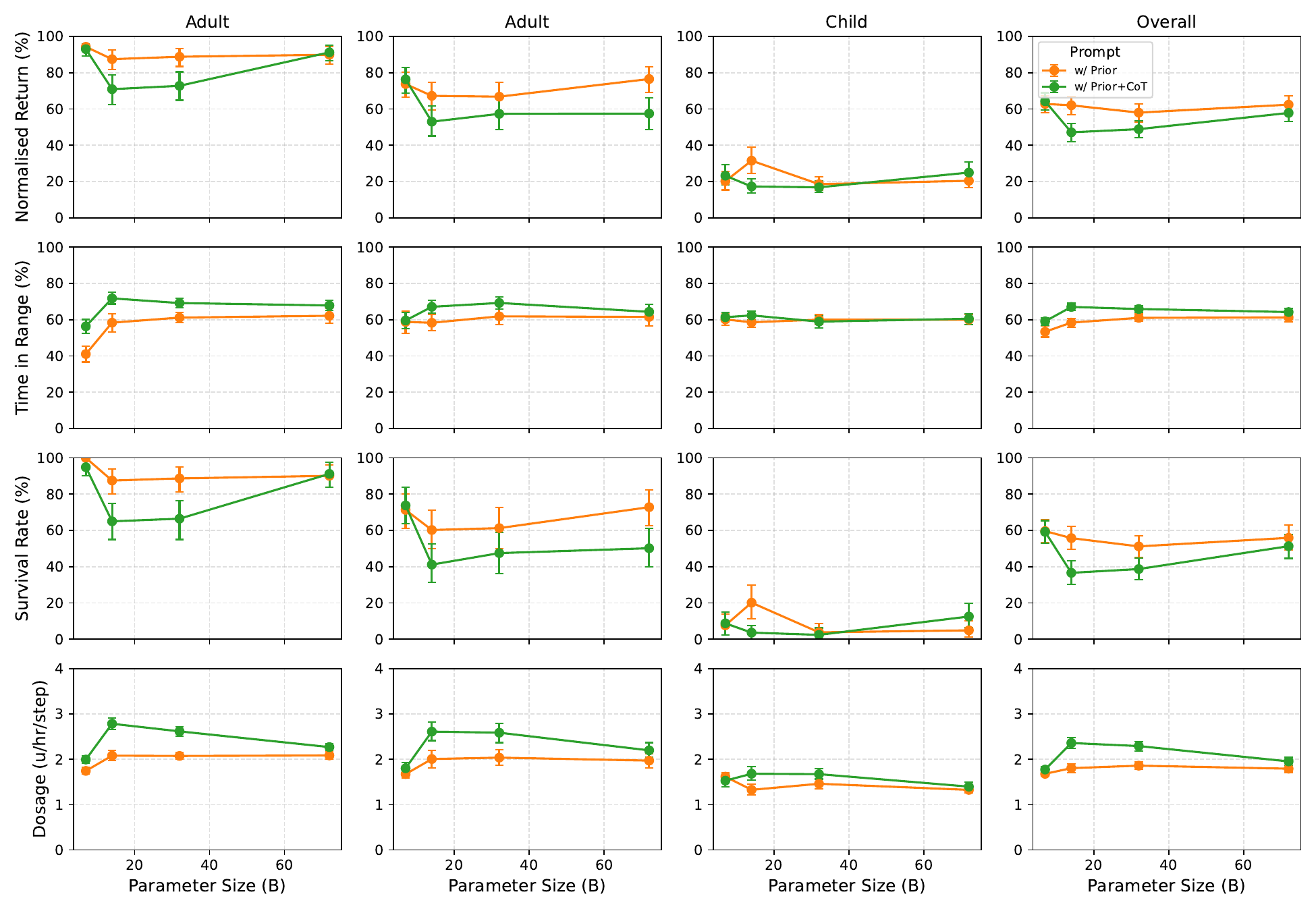}
    \caption[Detailed analysis of CoT prompting effects on insulin dosing performance within the Qwen2.5 family.]{Detailed analysis of CoT prompting effects on insulin dosing performance within the Qwen2.5 family. Temperature is set to 0.7. Each column corresponds to the evaluation result for adult (easy), adolescent (medium), child (hard), and overall, respectively. Each row of plots indicates a different evaluation metric, naming normalised return, TIR, survival rate and dosage (See Section~\ref{sec:llm-Reward and Evaluation Protocol}). The x-axis for each subplot indicates the parameter size of the Qwen2.5 model being reported. On the legend, 'w/ prior' means prior knowledge zero-shot prompt and 'w/ prior+CoT' means prior knowledge zero-shot-CoT prompt.}
    \label{fig:closer look at cot}
\end{figure}

Focussing exclusively on the Qwen2.5 family at a fixed decoding temperature of 0.7, we provide a detailed analysis of the nuanced effects of CoT prompting (Figure~\ref{fig:closer look at cot}). Shown in the last row of Figure~\ref{fig:closer look at cot}, explicit CoT prompts frequently result in overly aggressive insulin dosing strategies. Although this aggressiveness leads to improved time-in-range metrics during periods of stable glucose (second row in Figure~\ref{fig:closer look at cot}), it concurrently increases severe hypoglycaemic episodes, thus reducing overall survival rates and normalised returns (see the first and third rows in Figure~\ref{fig:closer look at cot}). Larger models mitigate these adverse effects through improved reasoning capabilities, suggesting that CoT prompts' efficacy critically depends on both task complexity and model scale. When comparing each column, it is shown that LLM tends to recommend less insulin to children than the other two strata.

\paragraph{Evaluating the Impact of Incorporating Meal Information.}
\begin{figure}
    \centering
    \includegraphics[width=0.9\linewidth]{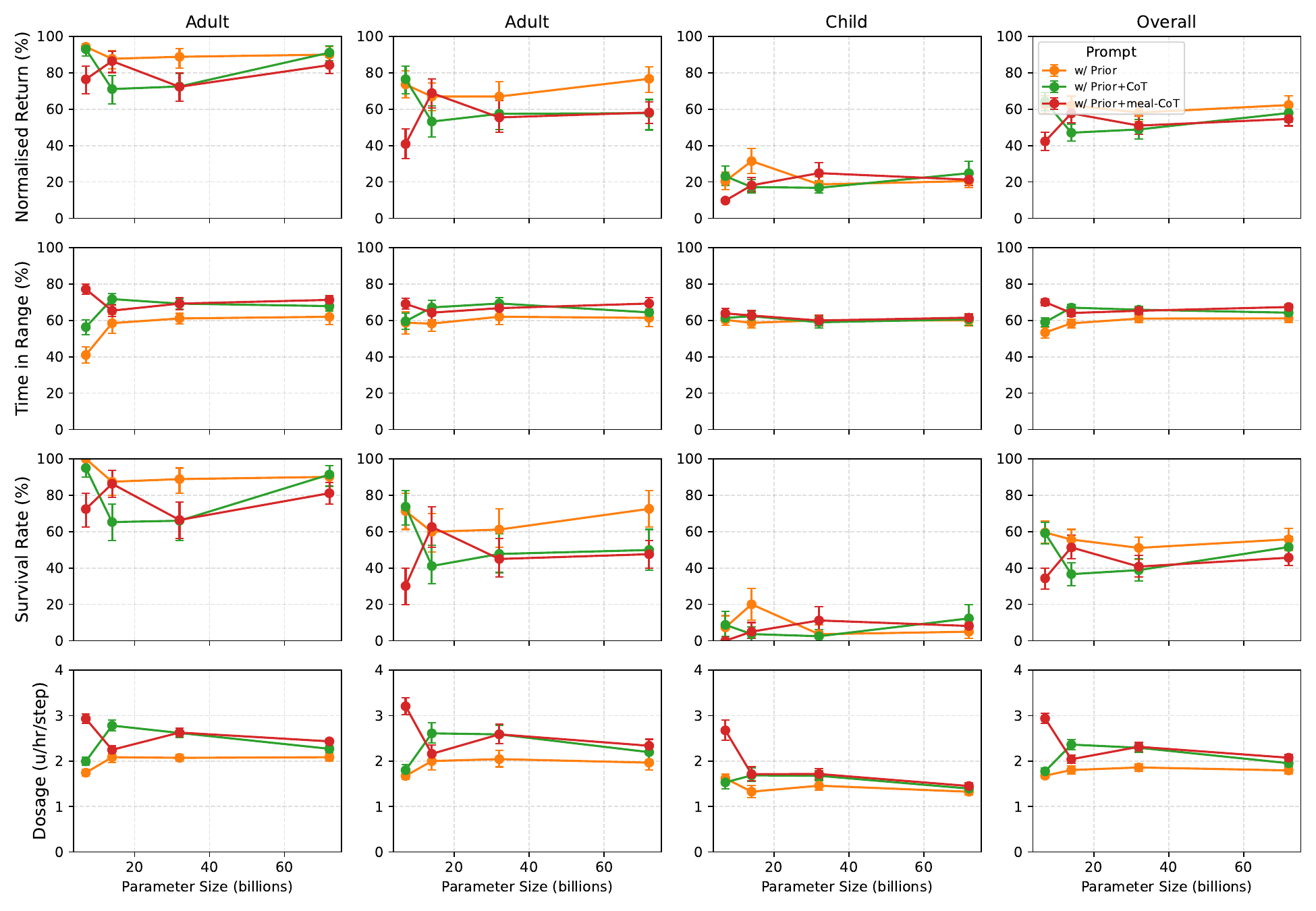}
    \caption[Effects of incorporating meal intake information into CoT prompts on insulin dosing performance.]{Effects of incorporating meal intake information into CoT prompts on insulin dosing performance. Temperature is set to 0.7. Each column corresponds to the evaluation result for adult, adolescent, child, and overall, respectively. Each row of plots indicates a different evaluation metric, naming normalised return, TIR, survival rate and dosage (See Section~\ref{sec:llm-Reward and Evaluation Protocol}). The x-axis for each subplot indicates the parameter size of the Qwen2.5 model being reported. On the legend, 'w/ prior' means prior knowledge zero-shot prompt, 'w/ prior+CoT' means prior knowledge zero-shot-CoT prompt and 'w. Prior+meal-cot' means the zero-shot CoT prompt with prior knowledge and meal information.}
    \label{fig:llm-meal-cot}
\end{figure}
To explore further improvements in clinical reasoning, we assess the inclusion of meal intake information in CoT prompting (Figure~\ref{fig:llm-meal-cot}). First, enriching the prompt with explicit meal-related reasoning (“Prior + meal-CoT”, red curves) consistently harms, rather than helps, the two smaller models (7 B and 13 B). Normalised return, time-in-range, and survival all decline relative to the simpler prior knowledge zero-shot baseline, while insulin dosage increases, indicating that the added reasoning chain is misinterpreted and translated into unnecessarily aggressive interventions. Secondly, the largest model (70 B) is able to absorb most of this complexity penalty. Its red curve returns to parity with the green (“w/ prior+CoT”) and orange baselines, but still fails to produce a systematic gain. In all capacities, there is no metric for which meal-based prompting delivers a statistically significant improvement.

Overall, the analysis indicates that while larger LLMs can ‘undo’ the performance drop caused by the specified meal reasoning prompt, they do not convert the hint of hidden variables into better glycaemic control. Therefore, latent-state inference of meal effects remains an unsolved challenge, and future work will need to look beyond prompt engineering to realise further clinical gains.

\paragraph{Behavioural Analysis of Model Responses}
\begin{figure}[htbp]
    \centering
    \includegraphics[width=0.99\textwidth]{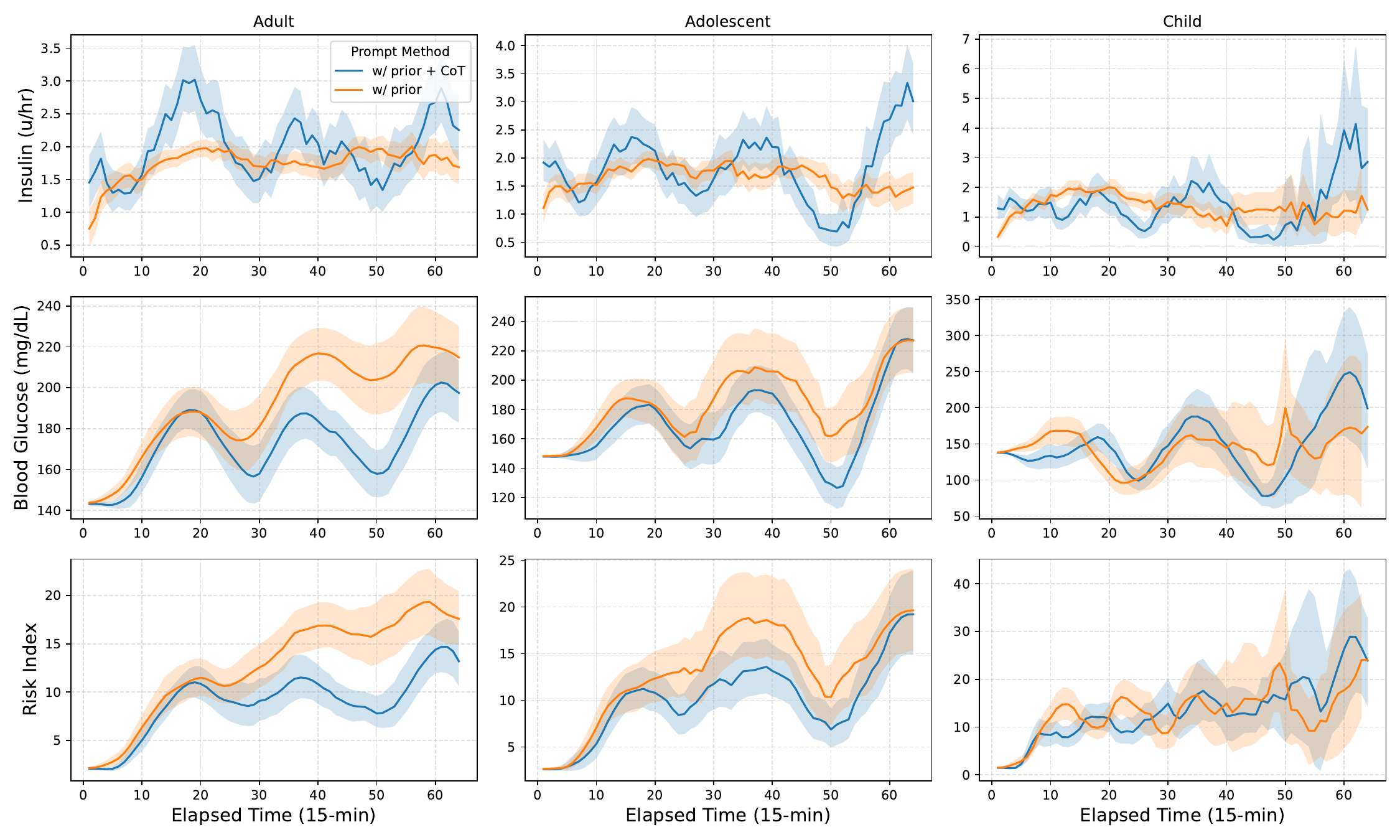}
    \caption[Insulin administration, glucose and risk behaviours for Qwen2.5-7B model under CoT and zero-shot prompting with prior knowledge system prompt.]{Insulin administration, glucose and risk behaviours for Qwen2.5-7B model under CoT and zero-shot prompting with prior knowledge system prompt. The shaded areas mean the 95\% confidence interval derived by bootstrapping. The x-axis is the elapsed time step. Each column corresponds to the evaluation result for adult, adolescent and child, respectively. The shaded areas mean the 95\% confidence interval derived by bootstrapping.}
    \label{fig:llm-behaviour-7b-1}
    
    \includegraphics[width=0.99\textwidth]{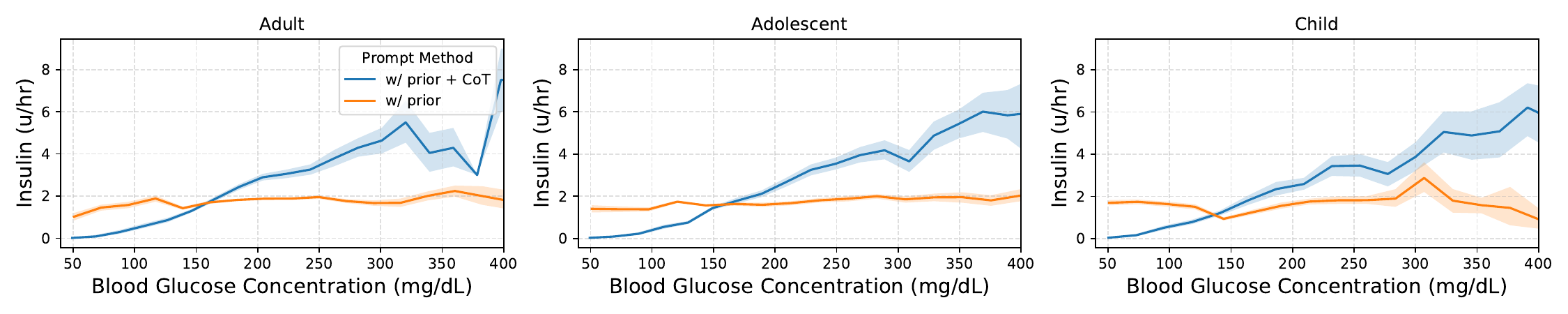}
    \caption[Insulin administration as a function of current observed glucose measurement for Qwen2.5-7B model under CoT and zero-shot prompting with prior knowledge system prompt.]{Insulin administration as a function of current observed glucose measurement for Qwen2.5-7B model under CoT and zero-shot prompting with prior knowledge system prompt. Notation is the same as Figure~\ref{fig:llm-behaviour-7b-1}.}
    \label{fig:llm-behaviour-7b-2}
\end{figure}

\begin{figure}[htbp]
    \centering
    \includegraphics[width=0.99\textwidth]{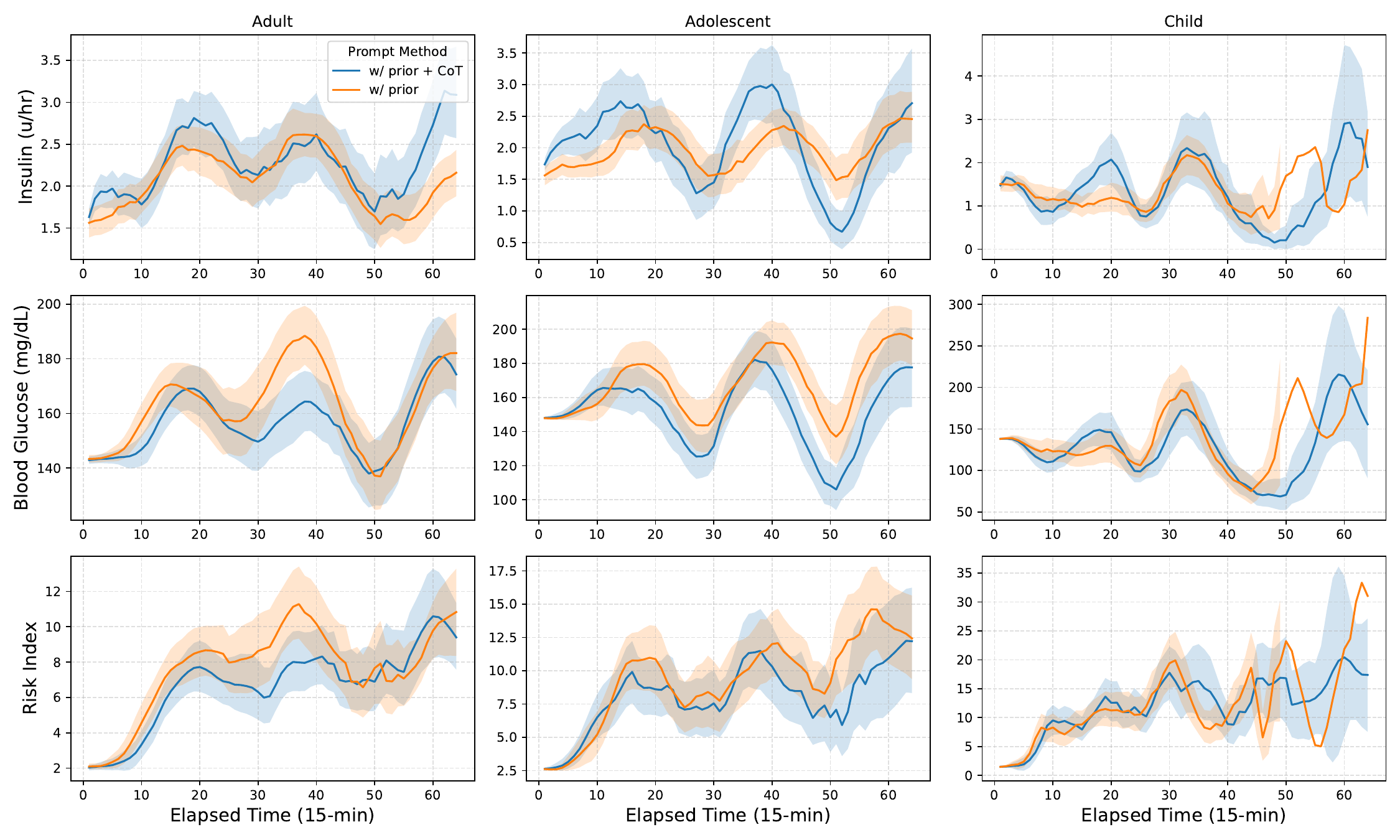}
    \caption[Insulin administration, glucose and risk behaviours for Qwen2.5-72B model under CoT and zero-shot prompting with prior knowledge system prompt.]{Insulin administration, glucose and risk behaviours for Qwen2.5-72B model under CoT and zero-shot prompting with prior knowledge system prompt. The shaded areas mean the 95\% confidence interval derived by bootstrapping. The x-axis is the elapsed time step. Each column corresponds to the evaluation result for adult, adolescent and child, respectively. The shaded areas mean the 95\% confidence interval derived by bootstrapping. When the shaded area does not show, it means the sample size is too small to present the confidence interval. The mean is used without a confidence interval in such cases.}
    \label{fig:llm-behaviour-72b-1}
    
    \includegraphics[width=0.99\textwidth]{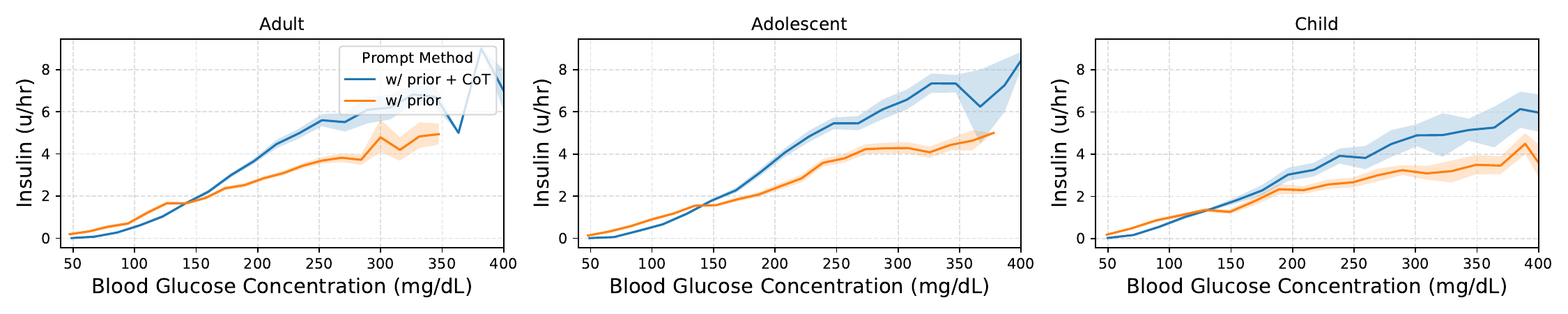}
    \caption[Insulin administration as a function of current observed glucose measurement for Qwen2.5-72B model under CoT and zero-shot prompting with prior knowledge system prompt.]{Insulin administration as a function of current observed glucose measurement for Qwen2.5-72B model under CoT and zero-shot prompting with prior knowledge system prompt. Notation is the same as Figure~\ref{fig:llm-behaviour-7b-1}}
    \label{fig:llm-behaviour-72b-2}
\end{figure}

We extend our analysis to behavioural patterns resulting from CoT prompting in the 7B (see Figure~\ref{fig:llm-behaviour-7b-1} and Figure~\ref{fig:llm-behaviour-7b-2}) and 72B (see Figure~\ref{fig:llm-behaviour-72b-1} and Figure~\ref{fig:llm-behaviour-72b-2}) parameter models. In the smaller 7B model, CoT prompts produce more reactive dosing strategies that dynamically adjust insulin rates according to glucose levels, resulting in improved short-term glucose control. However, these reactive policies also frequently lead to cumulative insulin overdoses, increasing hypoglycemic risk and reducing overall clinical outcomes. Specifically, the 7B model demonstrates an overly aggressive response to transient hyperglycaemic episodes, resulting in rapid insulin escalation and subsequent hypoglycaemic events. This behaviour indicates a limited ability to accurately estimate long-term insulin effects and highlights sensitivity to the instructional clarity of the prompt structure. In contrast, the larger 72B model inherently recognises glucose-dose relationships even in zero-shot scenarios, suggesting an intrinsic understanding of physiological insulin dynamics. The introduction of CoT further reinforces this understanding, allowing for a nuanced dosage adjustment more closely aligned with clinical guidelines. Critically, the increased complexity of CoT does not significantly increase hypoglycemic risks, demonstrating a robust enough model capacity to effectively balance aggressive glucose control against safety constraints.

\section{Failure Modes of Chain-of-Thought Reasoning}
 In this section, we analyse common failure modes of CoT reasoning when applied to insulin dosing in paediatric patients with Type 1 Diabetes. We focus on two illustrative case studies and supplement them with broader behavioural trends observed during the evaluation. The outputs of the Qwen2.5-7B and Qwen2.5-72B models are compared with the same prior knowledge and meal-CoT prompt.
 
\subsection{Case Study 1: Reasoning for Initial Dosage}
\label{sec:llm-case study1}
\begin{figure}[ht]
    \centering
    \includegraphics[width=\linewidth]{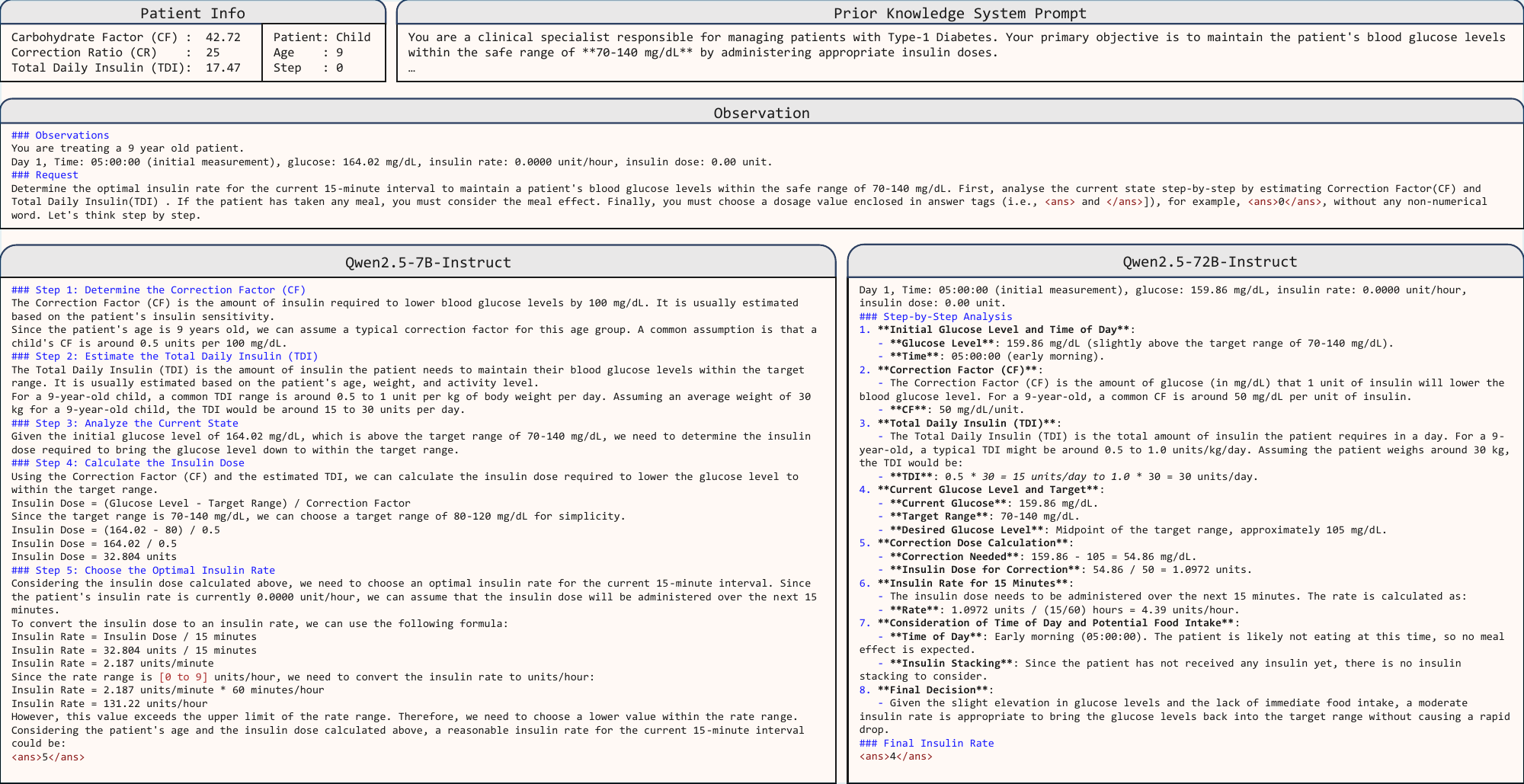}
    \caption[Comparing Qwen2.5 7B and 72B model response on a child patient on step 0 under meal CoT prompt with prior knowledge.]{Comparing Qwen2.5 7B and 72B model response on a child patient on step 0 under meal CoT prompt with prior knowledge. The system prompt and observation are identical, and responses are shown side by side.}
    \label{fig:llm-case1}
\end{figure}

We first examine a controlled single-step dosing scenario to evaluate how language models respond to an initial hyperglycaemic state with minimal context (see Figure~\ref{fig:llm-case1}). The virtual patient is a nine-year-old child whose initial glucose level at 05:00 is 164.02 mg/dL. Insulin has not been administered before this point, and no meal intake is recorded. The optimal strategy, derived from the ground truth of the simulator, involves calculating a correction dose based on an estimated correction factor of 25 mg/dL per unit and a total daily insulin of 17.47 units. This corresponds to a correction dose of approximately 2.36 units.

The Qwen2.5-7B model exhibits a clear reasoning failure. It begins by estimating an implausible CF of 0.5 units per 100 mg/dL and proceeds to subtract this from a target glucose level of 80 mg/dL. This leads to a highly exaggerated correction dose that exceeds 30 units. The model then calculates a per-hour dose exceeding 130 units/hour, which it subsequently clips to 5 units/hour in the final answer. Although the final output appears clinically plausible, it is numerically disconnected from the preceding reasoning trace. The model does not validate its recommendation against its own TDI estimate, nor does it apply any internal consistency constraints that would be expected in a clinical dose calculation. This illustrates a form of arithmetic hallucination, in which each individual step may appear locally coherent, but the overall logic is conceptually invalid and yields dangerous recommendations.

In contrast, the Qwen2.5-72B model produces a more restrained response. It estimates a CF of 50 mg/dL per unit, computes a correction dose of approximately 1.1 units, and derives an insulin rate of 4.39 units/hour, which is then rounded to 4.0. This outcome is physiologically reasonable and demonstrates a basic alignment with the underlying insulin-glucose dynamics. However, the model still underdoses the patient relative to the optimal dose derived from the simulator, reflecting a conservative bias when uncertainty exists. Although the internal calculations are more plausible than those of the smaller model, the final recommendation remains somewhat decoupled from the model's own TDI estimate and lacks any mechanism for deferred or conditional action.

\subsection{Case Study 2: Reasoning on Glucose Trends and Delayed Effects}
\begin{figure}[ht]
    \centering
    \includegraphics[width=\linewidth]{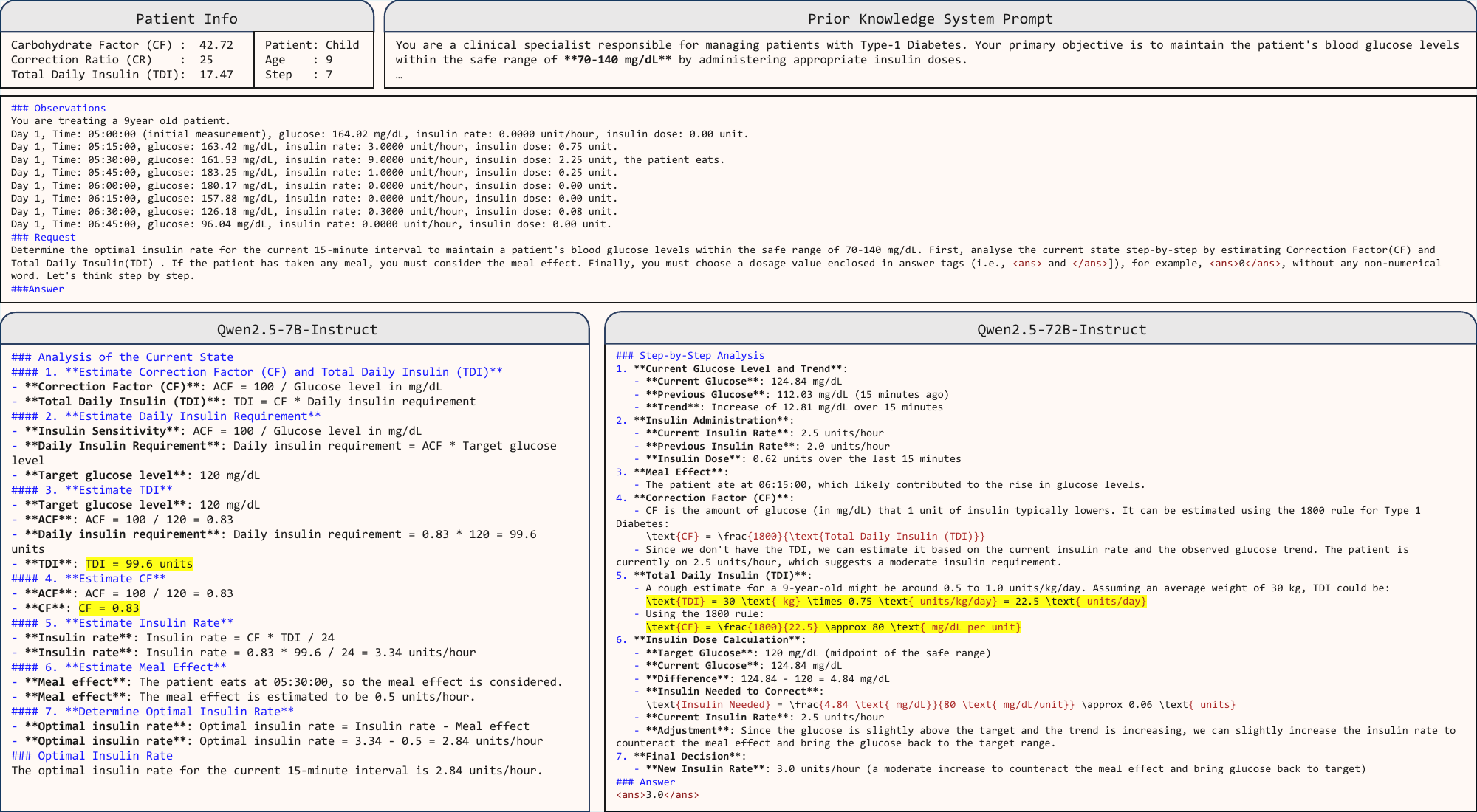}
    \caption[Comparing Qwen2.5 7B and 72B model response on a child patient on step 7 under meal CoT prompt with prior knowledge.]{Comparing Qwen2.5 7B and 72B model response on a child patient on step 7 under meal CoT prompt with prior knowledge. The system prompt and observation are identical, and responses are shown side by side. Texts highlighted in yellow denote wrong estimations.}
    \label{fig:llm-case2}
\end{figure}
The second scenario evaluates model behaviour in a temporally extended setting involving multiple insulin interventions, a recorded meal, and a non-linear glucose trend. The virtual patient is again the nine-year-old child reported in Section~\ref{sec:llm-case study1}. A meal event is recorded at 05:30, and insulin doses are administered at varying rates between 0.0 and 9.0 units/hour over the following six time steps.

The Qwen2.5-7B model fails to produce a coherent or physiologically plausible recommendation. It estimates the TDI using an incorrect and circular formula in which it multiplies a target glucose level by a fraction involving the same target value. This yields an inflated TDI of 99.6 units per day, approximately six times the true requirement for the patient. Based on this estimate, it derives a CF of 0.83 mg/dL per unit and calculates a correction dose of 3.34 units. Despite acknowledging the presence of a residual meal effect, the model merely subtracts a fixed offset of 0.5 units/hour and recommends a final insulin rate of 2.84 units/hour. This recommendation is aggressive and poorly justified, especially since the glucose value is within the safe range and trending downward. The model does not refer to or incorporate prior insulin administration, nor does it reason about possible insulin stacking or delayed pharmacodynamic effects.

The Qwen2.5-72B model performs better in terms of internal numerical consistency, but still fails to apply clinically appropriate decision logic. It estimates a TDI of 22.5 units and a CF of approximately 80 mg/dL per unit. It recognises that the glucose level has decreased over time after a meal, and it acknowledges that the current glucose level lies within the safe range. However, the model recommends increasing the insulin rate from 2.5 to 3.0 units/hour. This decision is based on a short-term increase in glucose between two adjacent time points, but it does not take into account the broader temporal context. Specifically, the recent increase in glucose is a transient fluctuation following a meal and is likely to self-correct due to existing insulin exposure. The model does not explicitly reason about cumulative insulin effects or the appropriate time lag between administration and glucose response.

\subsection{Summary of Findings on Case Studies}
Our case studies suggest that while CoT enables LLMs to simulate structured clinical reasoning, it also introduces characteristic errors. These errors often emerge not from model incapacity alone, but from the way CoT prompts compel deterministic, single-step answers in a domain that requires cautious, temporally-aware reasoning. To organise these insights, we group the observed failures into the following categories:

\begin{itemize}
    \item \textbf{Arithmetic Hallucination} The model performs invalid or circular arithmetic operations, often producing superficially coherent but numerically incorrect results. These errors typically arise from misapplied formulas or misinterpreted clinical heuristics during chain-of-thought reasoning.
    
    \item \textbf{Lack of Temporal Abstraction} The model fails to incorporate relevant treatment and meal history, such as insulin stacking or the delayed pharmacodynamic effects of meals and insulin. It tends to overreact to short-term glucose fluctuations while ignoring broader temporal trends, resulting in reactive rather than anticipatory control. This limitation reflects the absence of multi-step planning or memory in the decision-making process.
    
    \item \textbf{Aggressive Dosing Under Uncertainty} In early or ambiguous states, where glucose trends have not yet stabilised, the model frequently recommends excessive insulin doses. This behaviour increases the risk of hypoglycaemia and contrasts with human clinical intuition, which generally favours cautious observation before intervention under uncertainty.
    
    \item \textbf{Inconsistent Internal Reasoning} The model's intermediate reasoning steps often contain internal contradictions or fail to constrain the final output. For example, estimated total daily insulin requirements may not align with the recommended dose, suggesting a lack of internal logical coherence across the reasoning trace.
\end{itemize}

\section{Discussion}
This work provides an exploratory but rigorous evaluation of LLMs as DTR agents for insulin administration in Type 1 diabetes within an in silico environment. Our findings provide several insights critical to assessing the potential of LLMs in clinical decision-making. Firstly, our experiments demonstrate that certain smaller-scale LLMs, notably the Qwen2.5-7B model, can achieve performance comparable to extensively trained RL agents under zero-shot inference. This highlights a remarkable efficiency advantage of LLMs, as they do not require interaction-intensive, environment-specific training, positioning them as promising candidates for rapid deployment in clinical decision-support scenarios. Interestingly, our results do not confirm a consistent parameter-scaling law in clinical DTR tasks despite the expectations of the broader literature on LLMs. Larger-scale models, such as the Qwen2.5 7B, did not produce significant performance improvements over their smaller counterparts. Specifically, models from the Qwen2.5 family consistently outperformed those from the LLaMA3 family, indicating that the pre-training quality of foundation models has a substantial influence on clinical decision-making capabilities.

Our investigation of prior knowledge injection strategies revealed contrasting outcomes depending on the underlying RL algorithm. For off-policy algorithms (DQN), embedding prior knowledge significantly improved performance in stable environments but negatively impacted volatile conditions, illustrating the context-sensitive nature of knowledge incorporation. In contrast, PPO, an on-policy algorithm, benefited modestly but consistently across environments from implicit expert-guided exploration. These results highlight the complexity of effectively embedding clinical heuristics in RL agents, reinforcing the potential advantage of LLMs, which naturally incorporate prior knowledge through intuitive natural language prompting.

A particularly unexpected finding refers to the effectiveness of the Chain-of-Thought prompting method. Unlike general-purpose tasks, where CoT consistently improves reasoning performance, in insulin dosing scenarios, CoT prompts led to mixed outcomes. For smaller models, CoT induced an overly aggressive insulin dosing, which improved short-term glucose control at the expense of increased hypoglycemic episodes, thus reducing overall clinical efficacy. Larger models mitigated but did not eliminate these adverse effects, indicating that CoT prompts, while beneficial for human readability, must be carefully tuned to clinical contexts to avoid unintended risks.

Efforts to explicitly incorporate latent variables such as meal intake into LLM reasoning yielded minimal additional benefit, suggesting a current limitation in LLMs' ability to infer complex, partially observable physiological dynamics purely through text-based prompts. This finding highlights a critical area for future research: the potential integration of explicit physiological models or structured hybrid prompting techniques to further enhance the interpretative capabilities of LLMs.

Finally, our detailed case study of failure modes in CoT reasoning emphasises the risks associated with relying solely on linguistic coherence and structured logic. Despite producing fluent and syntactically well-formed reasoning steps, LLMs often exhibited fundamental clinical errors, such as misinterpreting glucose trends, overestimating insulin requirements, or failing to account for delayed pharmacodynamic effects. The contrast between LLM-generated decisions and standard clinical practice, where deferred intervention is often a safer and more appropriate choice, highlights a critical gap in the temporal and probabilistic reasoning capabilities of current LLMs. This limitation underscores the need for further research into prompt designs that support multi-step planning, uncertainty awareness, and dynamic policy modulation based on evolving patient states.

In summary, our investigation highlights the substantial potential of LLM as interpretable, efficient, and scalable agents for clinical decision support in dynamic treatment scenarios. However, realising this potential requires meticulous validation, context-sensitive prompt engineering, and potentially targeted fine-tuning. These findings advocate for a cautious, yet optimistic outlook on integrating LLMs into clinical workflows, highlighting the need for rigorous empirical evaluations to ensure safety and efficacy in practical healthcare settings.

\bibliographystyle{unsrt}  
\bibliography{references}

\appendix

\input{appendix/appendix-llm}
\end{document}

%% file: appendix/appendix-llm.tex
\section{Description for State Space, Action Space and Reward}
\label{app:sec-state-action-reward}
The \textit{SimGlucoseEnv}~\cite{luo2024dtr} simulates glucose-insulin dynamics to evaluate insulin dosing strategies in managing Type 1 diabetes. The environment’s state space primarily includes plasma glucose concentrations measured in mg/dL, historical glucose trajectories, insulin infusion histories, and indicators of meal intake.

The action space represents continuous insulin dosing rates, ranging between 0 and 9 units/hour, administered every 15 minutes. Insulin doses are evenly distributed throughout the subsequent 15-minute interval.

The reward function is specifically constructed to encourage robust glycaemic control while mitigating clinical risks. The risk component assesses deviations from clinically optimal glucose ranges, imposing penalties for hypoglycaemia and hyperglycaemia episodes. Episodes terminate prematurely upon encountering extreme glucose levels (below 40 mg/dL or above 500 mg/dL), incurring a substantial penalty to strongly discourage unsafe dosing behaviours. Episodes that remain within these safe glucose boundaries continue for a maximum duration of 64 time steps (16 hours), after which they are considered truncated without additional reward or penalty. This reward formulation explicitly balances effective glucose management against patient safety constraints.

The reward $r_t$ at time step $t$ is formulated as follows:

$$
r_t = r_{\text{termination}} + r_{\text{risk}}
$$

The termination reward, $r_{\text{termination}}$, explicitly penalises episodes terminated due to extreme plasma glucose levels:

$$
r_{\text{termination}} = 
\begin{cases}
-100, & \text{if terminated due to extreme BG (<40 mg/dL or >500 mg/dL)},\\[6pt]
0, & \text{otherwise (including truncated episodes)}.
\end{cases}
$$

The risk-based reward component, $r_{\text{risk}}$, is computed using the glucose risk index (RI), defined as:

$$
\text{RI}(t) = \text{LBGI}(t) + \text{HBGI}(t),
$$

where the Low Blood Glucose Index (LBGI) and High Blood Glucose Index (HBGI) are derived from a transformed glucose value $f_{\text{BG}}(t)$:

$$
f_{\text{BG}}(t) = 1.509 \cdot \left(\ln(\text{BG}(t))^{1.084} - 5.381\right),
$$

with LBGI and HBGI calculated as:

$$
\text{LBGI}(t) = 10 \cdot \text{mean}\left(f_{\text{BG}}(t)^2 \mid f_{\text{BG}}(t) < 0\right),
$$

$$
\text{HBGI}(t) = 10 \cdot \text{mean}\left(f_{\text{BG}}(t)^2 \mid f_{\text{BG}}(t) > 0\right).
$$

The risk-based reward is scaled linearly to the interval $[-100, 0]$:

$$
r_{\text{risk}}(t) = \frac{-\text{RI}(t) - (-100)}{0 - (-100)}.
$$

\section{Action Transformation in Continuous Action Space: A Comparative Analysis of clip and tanh}
\label{app:llm-tanh}
In continuous control problems with bounded action spaces, reinforcement learning (RL) policy networks must transform unbounded latent logits $z \in \mathbb{R}^d$ into bounded physical actions $a \in [0, D_{\max}]^d$. This is commonly achieved using a two-step transformation:

$$
z \sim \mathcal{N}(\mu_\theta(s), \operatorname{diag}(\sigma_\theta^2(s))), \quad a = \frac{D_{\max}}{2}\left(f(z) + \mathbf{1}\right),
$$

where $f: \mathbb{R}^d \to [-1, 1]^d$ is an elementwise squashing function. Two common functions $f$ are:

$$
f(z) = \begin{cases}
\tanh(z), & \text{(smooth squash)},\\[4pt]
\operatorname{clip}(z, -1, 1), & \text{(hard cut)}.
\end{cases}
$$

\subsection{Theoretical Properties of Tanh Transformation}

The tanh function is differentiable and invertible from $\mathbb{R} \to (-1,1)$, allowing precise computation of transformed action probabilities through the change-of-variables formula. Given:

$$
a = g(z) = \frac{D_{\max}}{2}(\tanh z + 1), \quad z = g^{-1}(a) = \operatorname{artanh}\left(\frac{2a}{D_{\max}} - 1\right),
$$

the log-density of the transformed action is:

$$
\log \pi_\theta(a \mid s) = \log \phi(z; \mu_\theta(s), \sigma_\theta(s)) - \sum_{i=1}^d \log\left(\frac{D_{\max}}{2}(1 - \tanh^2 z_i)\right).
$$

This approach maintains consistency in sampling and probability evaluation, preserving the validity of the PPO likelihood ratio and KL divergence computations. Additionally, the tanh Jacobian term penalises extreme logits, implicitly favouring safer, smaller interventions unless explicitly justified by strong model evidence.

\subsection{Theoretical Limitations of Clip Transformation}

In contrast, the clip function $f(z) = \operatorname{clip}(z, -1, 1)$ is neither differentiable nor invertible outside $[-1,1]^d$. The correct action probability distribution under clipping would need to be represented as a mixture distribution, combining truncated Gaussian densities within bounds and discrete probability mass at boundary points. However, most implementations incorrectly treat clipped actions as directly sampled from a Gaussian distribution without accounting for clipping effects, leading to several critical issues:

\begin{itemize}
\item \textbf{Incorrect likelihood ratios}: Clipped actions mismatch actual sampling distributions, invalidating the PPO theoretical assumptions.
\item \textbf{Misleading KL divergence}: The computed KL divergence ignores action saturation, failing to correctly constrain boundary behaviours.
\item \textbf{Gradient instability}: Non-differentiable clipping results in zero or incorrect gradients, reducing policy training effectiveness.
\end{itemize}

\subsection{Practical Considerations}
Despite the clip's theoretical limitations, clipping remains widely used due to its simplicity and numerical stability. While this may be acceptable for general-purpose RL tasks without stringent safety requirements, we argue that the potential clinical risks and theoretical inconsistencies could make clipping unsuitable for medical decision-making tasks. Therefore, we choose to adopt the tanh-based transformation for RL applications in healthcare settings as expert knowledge.

\section{Heuristic Decision for Exploration Strategy in Discrete Action Space}
\label{app:llm-heuristic}

In reinforcement learning settings involving clinical dosing, naïve uniform exploration often results in implausible or unsafe actions, particularly when the optimal policy frequently prescribes no intervention. To mitigate this, we introduce a domain-informed variant of $\epsilon$-greedy exploration that assigns disproportionately high probability to the zero-dose action, reflecting the real-world sparsity of insulin administration events.

Specifically, given a discretised action space of size $||\mathcal{A}||$, the probability of choosing action 0 (corresponding to a zero-dose) during exploration is defined as

$$
p_0 = \frac{||\mathcal{A}||(||\mathcal{A}||+1)}{2||\mathcal{A}||^2 - ||\mathcal{A}|| + 1},
$$

with the remaining probability mass $(1 - p_0)$ distributed uniformly across the $||\mathcal{A}||-1$ non-zero actions:

$$
p_j = \frac{1 - p_0}{||\mathcal{A}|| - 1}, \quad j = 1, \dots, ||\mathcal{A}||-1.
$$

This formulation ensures that the ratio between $p_0$ and any other $p_j$ scales linearly with $N$, thereby preserving a strong safety bias even as the dose grid becomes more finely discretised.

\begin{figure}
    \centering
    \includegraphics[width=0.9\linewidth]{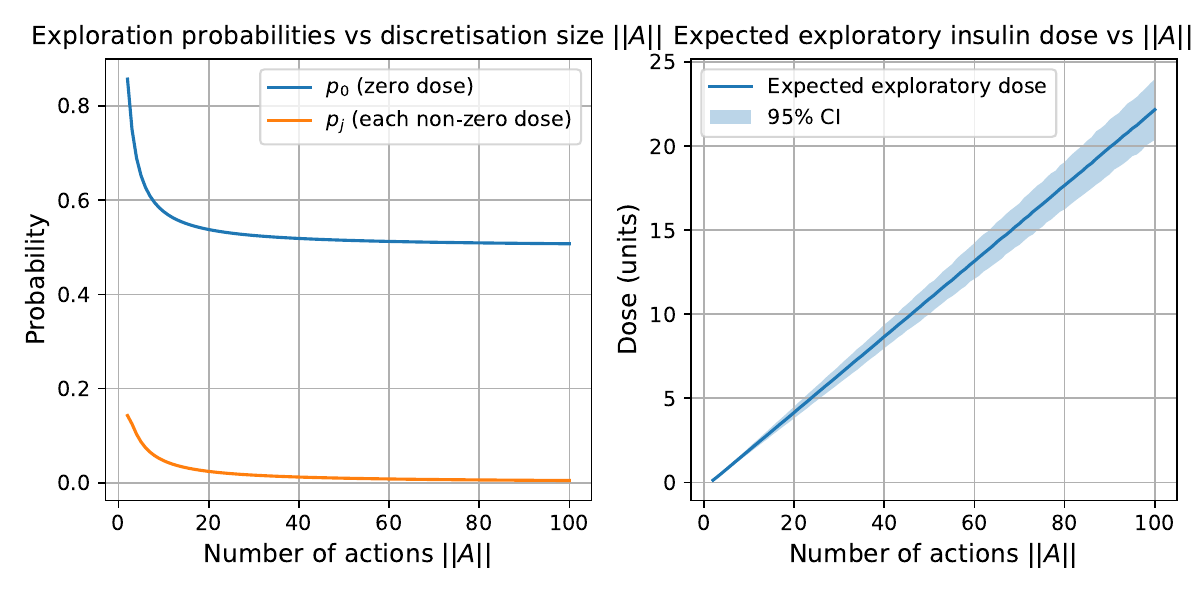}
    \caption[Exploration probabilities and expected exploratory insulin dose as a function of discretisation size $||\mathcal{A}||$.]{Exploration probabilities and expected exploratory insulin dose as a function of discretisation size $||\mathcal{A}||$. \textbf{Left}: Probability assigned to the zero-dose action and each non-zero action. As $||\mathcal{A}||$increases, it stabilises above 0.5. \textbf{Right}: Expected insulin dose under the exploration distribution, assuming non-zero actions correspond to dose units. The shaded region denotes the 95\% bootstrap confidence interval.}
    \label{fig:llm-exploration_probs}
\end{figure}.

Figure~\ref{fig:llm-exploration_probs} illustrates the evolution of $p_0$ and $p_j$ as a function of $||\mathcal{A}||$. The probability mass allocated to the zero-dose action quickly stabilises around 0.52 as $N$ increases, while the per-action mass for each non-zero dose diminishes accordingly. This ensures that, despite an expanding action space, the exploratory policy continues to prioritise conservative actions and the expectation of exploratory dose can grow linearly with the number of actions that increase.

\section{Hyperparameters for Small Reinforced Agent Training}
\label{app:llm-sra-hyperparameters}

\begin{table}[h]
\centering
\small
\caption{Common hyperparameters across all RL agents}
\label{tab:llm-common-hyperparams}
\begin{tabular}{ll}
\toprule
\textbf{Parameter} & \textbf{Value(s)} \\
\midrule
Seed & \{1, 100, 1000, 10000\} \\
Learning rate ($lr$) & \{1e-3, 1e-4\} \\
Batch size & 128 \\
Use expert knowledge & \{False, True\} \\
Observation mode & Stack(48) \\
Discount factor ($\gamma$) & 0.99 \\
Exploration noise & 0.1 \\
Test-time $\epsilon$ ($\epsilon_\text{test}$) & 0.001 \\
\bottomrule
\end{tabular}
\end{table}

\vspace{1em}

\begin{table}[h]
\centering
\small
\caption{DQN-specific hyperparameters}
\label{tab:llm-dqn-hyperparams}
\begin{tabular}{ll}
\toprule
\textbf{Parameter} & \textbf{Value(s)} \\
\midrule
Target network update frequency & \{100, 1000\} \\
Update per step & 1 \\
Actor update frequency & 1 \\
Train-time $\epsilon$ (initial) & 0.9 \\
Train-time $\epsilon$ (final) & 0.1 \\
$\epsilon$ schedule & Linear decay from 0.9 to 0.1 \\
\bottomrule
\end{tabular}
\end{table}

\vspace{1em}

\begin{table}[h]
\centering
\small
\caption{PPO-specific hyperparameters}
\label{tab:llm-ppo-hyperparams}
\begin{tabular}{ll}
\toprule
\textbf{Parameter} & \textbf{Value(s)} \\
\midrule
On-policy steps per collect & 192 \\
Repeat per collect & 20 \\
GAE lambda ($\lambda$) & 0.95 \\
Conditioned $\sigma$ & \{True, False\} \\
Value function coefficient ($v_\text{coef}$) & 0.5 \\
Entropy coefficient ($\text{ent}_\text{coef}$) & 0.001 \\
PPO clip range ($\epsilon_\text{clip}$) & 0.1 \\
Value function clipping & \{True, False\} \\
Dual clip & None \\
Advantage normalisation & True \\
Recompute advantage & False \\
\bottomrule
\end{tabular}
\end{table}